\documentclass{sig-alternate}

\usepackage{colortbl}
\usepackage{url}

\usepackage{tikz}
\usetikzlibrary{bayesnet}

\usepackage{color,soul}

\usepackage{url}
\usepackage[naturalnames,colorlinks,
      pdfpagemode={UseOutlines},
      bookmarks,
      bookmarksopen=true,
      pdfstartview={FitH},
      bookmarksnumbered=true,
      linkcolor=blue,
            citecolor=blue,
            urlcolor=magenta,
            linktocpage,
            plainpages=false,
        pdfauthor={Siddharth Reddy, Igor Labutov, Thorsten Joachims},
      pdftitle={Latent Skill Embedding for Personalized Lesson Sequence Recommendation},
            pdftex
]{hyperref}

\begin{document}






%

\title{Latent Skill Embedding for Personalized Lesson Sequence Recommendation}
%
%
%
%
%

\numberofauthors{3} 
%
\author{
%
%
\alignauthor
Siddharth Reddy\\
       \affaddr{Department of Computer Science}\\
       \affaddr{Cornell University}\\
       \affaddr{Ithaca, NY}\\
       \email{sgr45@cornell.edu}
\alignauthor
Igor Labutov\\
       \affaddr{Department of Electrical and Computer Engineering}\\
       \affaddr{Cornell University}\\
       \affaddr{Ithaca, NY}\\
       \email{iil4@cornell.edu}
\alignauthor Thorsten Joachims\\
       \affaddr{Department of Computer Science}\\
       \affaddr{Cornell University}\\
       \affaddr{Ithaca, NY}\\
       \email{tj@cs.cornell.edu}
}

\maketitle
\begin{abstract}
Students in online courses generate large amounts of data that can be used to personalize the learning process and improve quality of education. In this paper, we present the Latent Skill Embedding (LSE), a probabilistic model of students and educational content that can be used to recommend personalized sequences of lessons with the goal of helping students prepare for specific assessments. Akin to collaborative filtering for recommender systems, the algorithm does not require students or content to be described by features, but it learns a representation using access traces. We formulate this problem as a regularized maximum-likelihood embedding of students, lessons, and assessments from historical student-content interactions. An empirical evaluation on large-scale data from Knewton, an adaptive learning technology company, shows that this approach predicts assessment results competitively with benchmark models and is able to discriminate between lesson sequences that lead to mastery and failure.
\end{abstract}

%
%


%
%

%
%



\section{Introduction}
The popularity of online education platforms has soared in recent years. Companies like Coursera and EdX offer Massive Open Online Courses (MOOCs) that attract millions of students and high-calibre instructors. Khan Academy has become a hugely popular repository of videos and interactive materials on a wide range of subjects. E-learning products offered by universities and textbook publishers are also gaining traction. These platforms improve access to high quality educational content for anyone connected to the Internet. As a result, people who would otherwise lack the opportunity are able to consume materials like video lectures and problem sets from courses offered at top universities. However, in these online environments learners often lack the personalized instruction and coaching that can potentially lead to significant improvements in educational outcomes. Furthermore, the educational content may be contributed by many authors without a formal underlying structure. Intelligent systems that learn about the educational properties of the content, guide learners through custom lesson plans, and quickly adapt through feedback could help learners take advantage of large and heterogeneous collections of educational content to achieve their goals.

The extensive literature on intelligent tutoring systems (ITS) and computer-assisted instruction (CAI) dates back to the 1960s. Early efforts focused on approximating the behavior of a human tutor through rule-based systems that taught students South American geography \cite{carbonell1970ai}, electronics troubleshooting \cite{lesgold1988sherlock}, and programming in Lisp \cite{CorbettBKT}. Today's online education platforms differ from early ITSes in their ability to gather data at scale, which facilitates the use of machine learning techniques to improve the educational experience. Relatively little academic work has been done to design systems that use the massive amounts of data generated by students in online courses to provide personalized learning tools. Learning and content analytics \cite{Lan:2014:SFA:2627435.2670314}, instructional scaffolding in educational games \cite{o2015framework}, hint generation \cite{piechetal15}, and feedback propagation \cite{piech2015learning} are a few topics currently being explored in the personalized learning space.

Our aim is to build a domain-agnostic framework for modeling students and content that can be used in many online learning products for personalized lesson sequence recommendation. A common data source available in products is a stream of interaction data, or \emph{access traces} that log student interactions with modules of course content. These access traces have the form \emph{Student A completed Lesson B} and \emph{Student C passed assessment D}. \emph{Lessons} are content modules that introduce or reinforce concepts; for example, an animation of cellular respiration or a paragraph of text on Newton's first law of motion. \emph{Assessments} are content modules with pass-fail results that test student skills; for example, a true-or-false question halfway through a video lecture. By relying on a coarse-grained, binary assessment result, we are able to gracefully handle many types of assessments (e.g., free response and multiple choice) as long as a student response can be labelled as correct or incorrect.

We use access traces to embed students, lessons, and assessments together in a joint semantic space, yielding a representation that can be used to reason about the relationship between students and content (e.g., the likelihood of passing an assessment, or the skill gains achieved by completing a lesson). The model is evaluated on simple synthetic scenarios, as well as large-scale real data from Knewton, an education technology company that offers personalized recommendations and activity analytics for online courses \cite{knewton2015}. The data set consists of 2.18 million access traces from over 7,000 students, recorded in 1,939 classrooms over a combined period of 5 months.

\section{Related Work}

Our work builds on the existing literature in psychometric user modeling. The Rasch model estimates the probability of a student passing an assessment using latent concept proficiency and assessment difficulty parameters \cite{rasch1993probabilistic}. The two-parameter logistic item response theory (2PL IRT) model adds an assessment discriminability parameter to the result likelihood \cite{linden1997handbook}. Both models assume that a map from assessments to a small number of underlying concepts is known a priori. We propose a data-driven method of learning content representation that does not require a priori knowledge of content-to-concept mapping. Though this approach sacrifices the interpretability of expert ratings, it has two advantages: 1) it does not require labor-intensive expert annotation of content and 2) it can evolve the representation over time as existing content is modified or new content is introduced.

Lan et al. propose a sparse factor analysis (SPARFA) approach to modeling graded learner responses that uses assessment-concept associations, concept proficiencies, and assessment difficulty \cite{Lan:2014:SFA:2627435.2670314}. The algorithm does not rely on an expert concept map, but instead learns assessment-concept associations from the data. Multi-dimensional item response theory \cite{m.d.reckase2009} also learns these assocations from the data. We extend the ideas behind SPARFA and multi-dimensional item response theory to include a model of student learning from lesson modules, which is a key prerequisite for recommending personalized lesson sequences.

Bayesian Knowledge Tracing (BKT) uses a Hidden Markov Model to model the evolution of student knowledge (which is discretized into a finite number of states) over time \cite{CorbettBKT}. Further work has modified the BKT framework to include the effects of lessons through an input-output Hidden Markov Model \cite{gonzalez2014general, pardos2012tutor, pardos2013adapting}. Similarly, SPARFA has been extended to model time-varying student knowledge and the effects of lesson modules \cite{lan2014time}. Item response theory has also been extended to capture temporal changes in student knowledge \cite{tskirt, jascha2014}. Recurrent neural networks have been used to trace student knowledge over time and model lesson effects \cite{DBLP:journals/corr/PiechSHGSGS15}. Similar ideas for estimating temporal student knowledge from binary-valued responses have appeared in the cognitive modeling literature \cite{prerau2008mixed, smith2004dynamic}. We extend this work in a multi-dimensional setting where student knowledge lies in a continuous state space and lesson prerequisites modulate knowledge gains from lesson modules.

Our model also builds on previous work that uses temporal embeddings to predict music playlists \cite{Moore/etal/13a}. While Moore et al. focused on embedding objects (songs) in a metric space, we propose a non-metric embedding where the distances between objects (students, assessments, and lessons) are not symmetric, capturing the natural progression in difficulty of assessments and the positive growth of student knowledge.

\section{Embedding Model}

We now describe the Latent Skill Embedding, a probabilistic model that places students, lessons, and assessments in a joint semantic space that we call the \emph{latent skill space}. Students have trajectories through the latent skill space, while assessments and lessons are placed at fixed locations. Formally, a student is represented as a set of $d$ latent skill levels $\vec{s} \in \mathbb{R}_+^d$; a lesson module is represented as a vector of skill gains $\vec{\ell} \in \mathbb{R}_+^d$ and a set of prerequisite skill requirements $\vec{q} \in \mathbb{R}_+^d$; an assessment module is represented as a set of skill requirements $\vec{a} \in \mathbb{R}_+^d$.

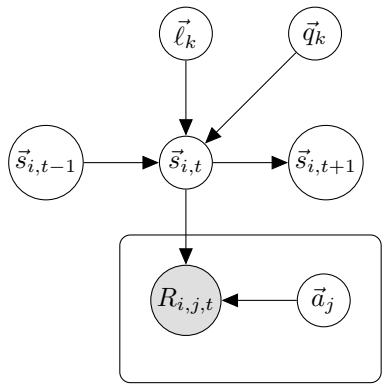
\begin{figure}
        \centering

        \begin{tikzpicture}

              \node[obs] (otcm) {$R_{i,j,t}$};
              \node[latent, above=of otcm] (scurr) {$\vec{s}_{i,t}$};
              \node[latent, above=of otcm, left=of scurr] (sprev) {$\vec{s}_{i,t-1}$};
              \node[latent, above=of scurr] (lesson) {$\vec{\ell}_k$};
              \node[latent, above=of scurr, right=of lesson] (prereq) {$\vec{q}_k$} ;
              \node[latent, below=of scurr, right=of otcm] (asmt) {$\vec{a}_j$};

              \node[latent, right=of scurr] (snext) {$\vec{s}_{i,t+1}$};

              \edge {sprev, lesson, prereq} {scurr} ;
              \edge {scurr, asmt} {otcm} ;

              \edge {scurr} {snext} ;

              \plate[xscale=1.2, yscale=1.4] {ao} {(asmt)(otcm)} {} ;

            \end{tikzpicture}

        \caption{A graphical model of student learning and testing, i.e. a continuous state space Hidden Markov Model with inputs and outputs. $\vec{s}$ = student knowledge state, $\vec{\ell}$ = lesson skill gains, $\vec{q}$ = lesson prerequisites, $\vec{a}$ = assessment requirements, and $R$ = result.}
        \label{fig:graphical-model}

\end{figure}

Students interact with lessons and assessments in the following way. First, a student can be tested on an assessment module with a pass-fail result $R \in \{0, 1\}$, where the likelihood of passing is high when a student has skill levels that exceed the assessment requirements and vice-versa. Second, a student can work on lesson modules to improve skill levels over time. To fully realize the skill gains associated with completing a lesson module, a student must satisfy prerequisites (only partly fulfilling the prerequisites will result in relatively smaller gains, see Equation \ref{eq:update-with-prereqs} for details). Time is discretized such that at every timestep $t \in \mathbb{N}$, a student completes a lesson and may complete zero or many assessments. The evolution of student knowledge can be formalized as the graphical model in Figure \ref{fig:graphical-model}, and the following subsections elaborate on the details of this model.

\begin{figure}
        \centering

        \includegraphics[width=\linewidth]{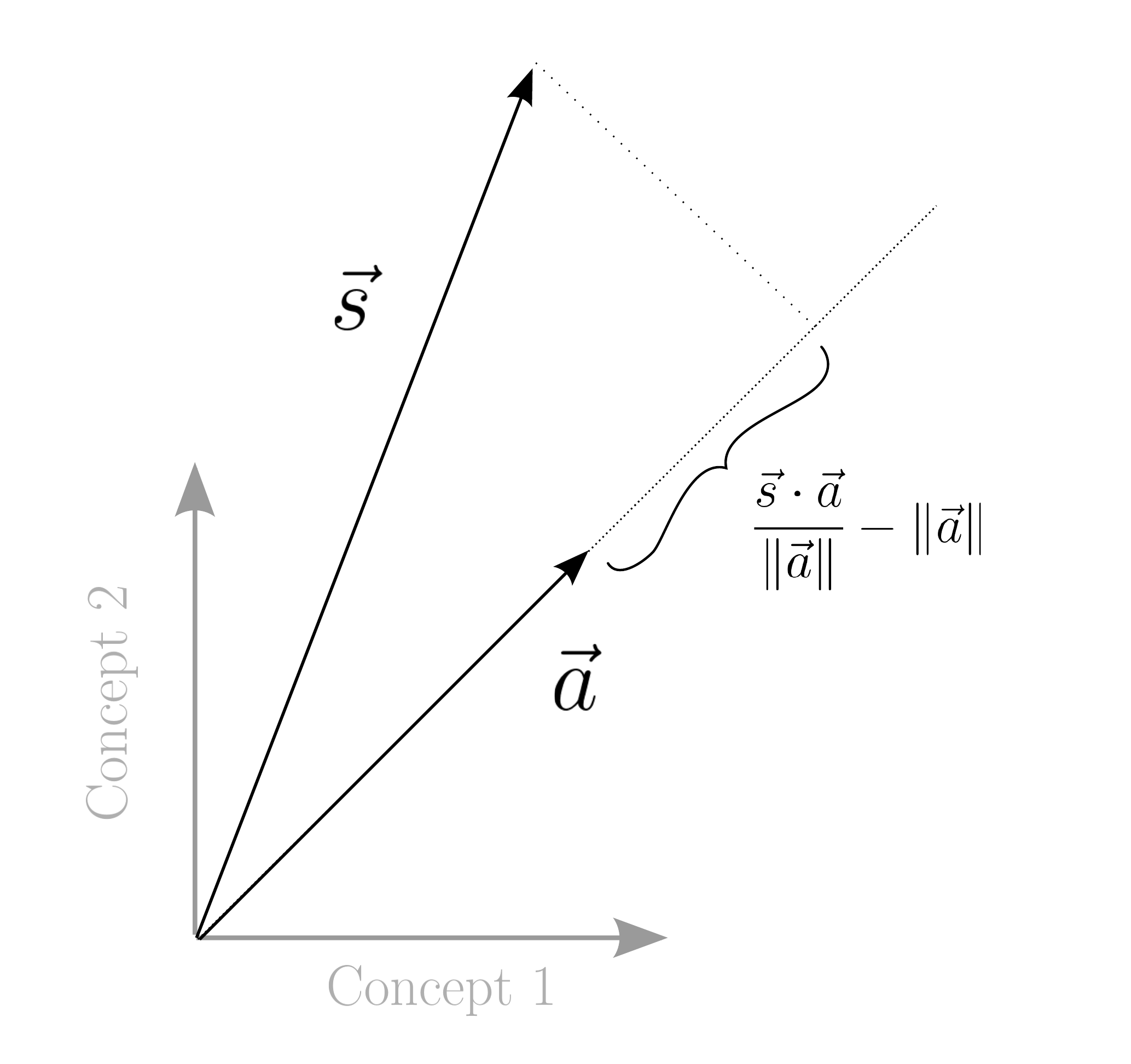}

        \caption{Geometric intuition underlying the parametrization of the assessment result likelihood (Equation \ref{eq:result-likelihood}). Only the length of the projection of the student's skills $\vec{s}$ onto the assessment vector $\vec{a}$ affects the pass likelihood of that assessment, meaning only the ``relevant'' skills (with respect to the assessment) should determine the result.}
        \label{fig:likelihood-geom}

\end{figure}

\subsection{Modeling Assessment Results}

For student $\vec{s}$, assessment $\vec{a}$, and result $R$,

\begin{equation} \label{eq:result-likelihood}
R \sim \mathrm{Bernoulli}(\phi(\Delta(\vec{s},~\vec{a})))
\end{equation}

where $\phi$ is the logistic function and $\Delta(\vec{s},~\vec{a}) = \frac{\vec{s} \cdot \vec{a}}{||\vec{a}||} - ||\vec{a}|| + \gamma_s + \gamma_a$. $\vec{s}$ and $\vec{a}$ are constrained to be non-negative (for details see the Parameter Estimation section). A pass result is indicated by $R=1$, and a fail by $R=0$. The term $\frac{\vec{s} \cdot \vec{a}}{||\vec{a}||}$ can be rewritten as $||\vec{s}||\mathrm{cos}(\theta)$, where $\theta$ is the angle between $\vec{s}$ and $\vec{a}$; it can be interpreted as ``relevant skill''. The term $||\vec{a}||$ can be interpreted as general (i.e. not concept-specific) assessment difficulty. The expression $\frac{\vec{s} \cdot \vec{a}}{||\vec{a}||} - ||\vec{a}||$ is visualized in Figure \ref{fig:likelihood-geom}. The bias term $\gamma_s$ is a student-specific term that captures a student's general (assessment-invariant and time-invariant) ability to pass; it can be interpreted as a measure of how well the student guesses correct answers. The bias term $\gamma_a$ is a module-specific term that captures an assessment's general (student-invariant and time-invariant) difficulty. $\gamma_a$ differs from the $||\vec{a}||$ difficulty term in that it is not bounded; see the Parameter Estimation section for details. These bias terms are analogous to the bias terms used for modeling song popularity in \cite{chen2012playlist}. Our choice of $\Delta$ differs from traditional multi-dimensional item response theory, which uses $\Delta(\vec{s}, \vec{a}) = \vec{s} \cdot \vec{a} + \gamma_a$ where $s$ and $a$ are not bounded (although in practice, suitable priors are imposed on these parameters).

\subsection{Modeling Student Learning from Lessons}

For student $\vec{s}$ who worked on a lesson with skill gains $\vec{\ell}$ and no prerequisites at time $t+1$, the updated student state is

\begin{equation} \label{eq:update-without-prereqs}
\vec{s}_{t+1} \sim \mathcal{N}\left(\vec{s}_t + \vec{\ell},~\Sigma\right)
\end{equation}

where the covariance matrix $\Sigma = I_d\sigma^2$ is diagonal. For a lesson with prerequisites $\vec{q}$,

\begin{equation} \label{eq:update-with-prereqs}
\vec{s}_{t+1} \sim \mathcal{N}\left(\vec{s}_t + \vec{\ell} \cdot \phi(\Delta(\vec{s}_t,~\vec{q})),~\Sigma\right)
\end{equation}

\begin{figure}
\centering
\includegraphics[width=\linewidth]{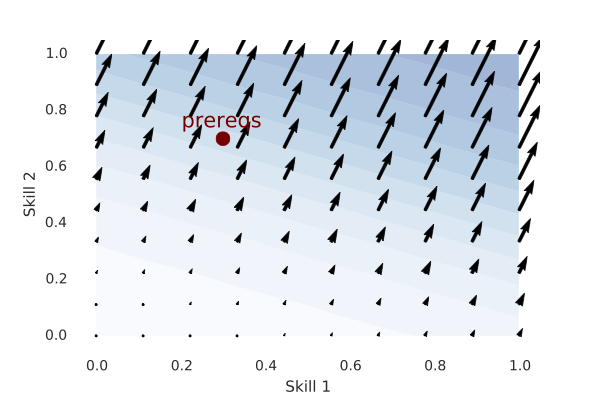}
\caption{The vector field of skill gains for a lesson with skill gains $\vec{\ell} = (0.5, 1)$ and prerequisites $\vec{q} = (0.7, 0.3)$. Contours are drawn for varying update magnitudes. A student can compensate for lack of prerequisites in one skill through excess strength in another skill, but the extent to which this trade-off is possible depends on the relative weights of the prerequisites.}
\label{fig:gain-vec-field}
\end{figure}

where $\phi$ is the logistic function and $\Delta(\vec{s}_t,~\vec{q}) = \frac{\vec{s}_t \cdot \vec{q}}{||\vec{q}||} - ||\vec{q}||$. The intuition behind this equation is that the skill gain from a lesson should be weighted according to how well a student satisfies the lesson prerequisites. A student can compensate for lack of prerequisites in one skill through excess strength in another skill, but the extent to which this trade-off is possible depends on the relative weights of the prerequisites. The same principle applies to satisfying assessment skill requirements. With prerequisites, the vector field of skill gains is non-uniform (without prerequisites, it is uniform); for example, see Figure \ref{fig:gain-vec-field}.

Our model differs from \cite{lan2014time} in that we explicitly model the effects of prerequisite knowledge on gains from lessons. Lan et al. model gains from a lesson as an affine transformation of the student's knowledge state.

\section{Parameter Estimation} \label{parameter-estimation}

We compute MAP estimates of model parameters $\Theta$ by maximizing the following objective function:

\begin{equation} \label{eq:opt-prob}
\begin{split}
L(\Theta) = \sum\limits_{\mathcal{A}} \log{(\mathbb{P}[R \mid \vec{s}_t, \vec{a}, \gamma_s, \gamma_a])} \\
+ \sum\limits_{\mathcal{L}} \log{(\mathbb{P}[\vec{s}_{t+1} \mid \vec{s}_t, \vec{\ell}, \vec{q}])} - \beta \cdot \lambda(\Theta)
\end{split}
\end{equation}

where $\mathcal{A}$ is the set of assessment interactions, $\mathcal{L}$ is the set of lesson interactions, $\lambda(\Theta)$ is a regularization term that penalizes the $L_2$ norms of embedding parameters (not bias terms), and $\beta$ is a regularization parameter. Non-negativity constraints on embedding parameters (not bias terms) are enforced.

$L_2$ regularization is used to penalize the size of embedding parameters to prevent overfitting. The bias terms are not bounded or regularized. This allows $-||\vec{a}|| + \gamma_a$ to be positive for assessment modules that are especially easy, and $\frac{\vec{s} \cdot \vec{a}}{||\vec{a}||} + \gamma_s$ to be negative for students who fail especially often. We solve the optimization problem with box constraints using the L-BFGS-B \cite{lbfgs} algorithm. We randomly initialize parameters and run the iterative optimization until the relative difference between consecutive objective function evaluations is less than $10^{-3}$. Averaging validation accuracy over multiple runs during cross-validation reduces sensitivity to the random initializations (since the objective function is non-convex).

\section{Experiments on Synthetic Data}

To verify the correctness of our model and to illustrate the properties of the embedding geometry that the model captures, we conducted a series of experiments on small, synthetically-generated interaction histories. Each scenario is intended to demonstrate a different feature of the model (e.g., recovering student knowledge and assessment requirements in the absence of lessons, or recovering sensible skill gain vectors for different lessons). For the sake of simplicity, the embeddings do not use bias terms. The scenarios shown next are annotated versions of plots made by our embedding software.

    \begin{figure}[t]
        \centering
        \includegraphics[width=\linewidth]{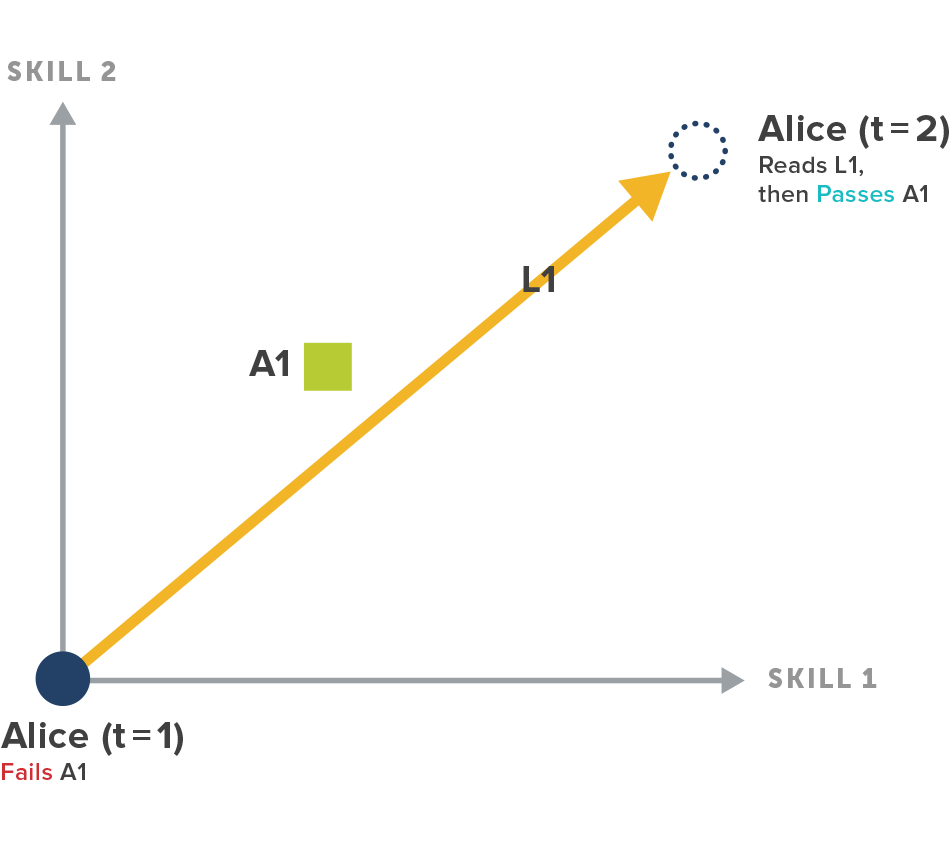}
        \caption{An extremely simple embedding}
        \label{fig:ex1}
    \end{figure}

    \begin{figure}[t]
        \centering
        \includegraphics[width=\linewidth]{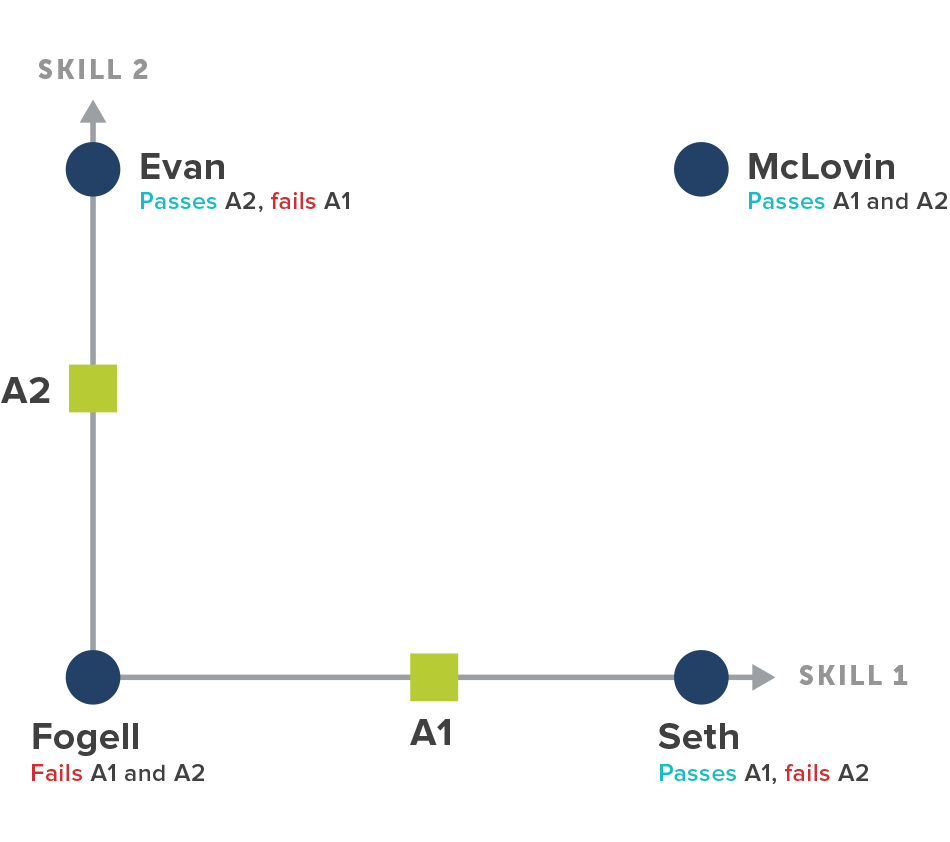}
        \caption{A two-dimensional embedding without lessons}
        \label{fig:ex2}
    \end{figure}

    \begin{figure}[t]
        \centering
        \includegraphics[width=\linewidth]{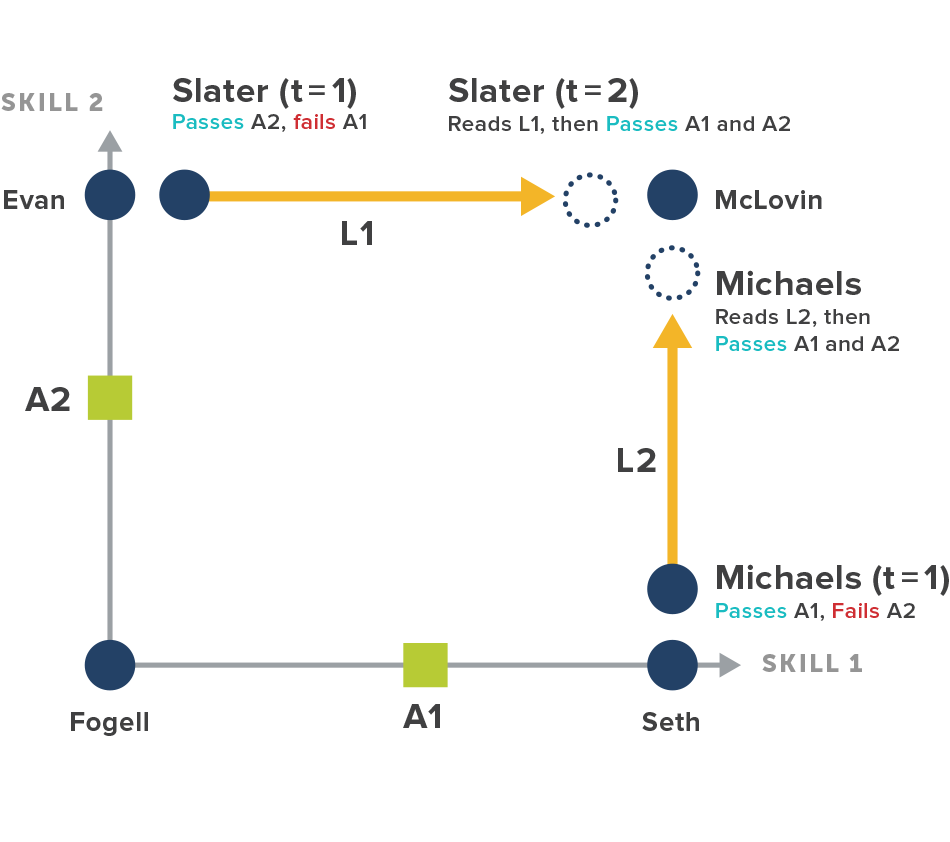}
        \caption{A two-dimensional embedding with lessons, without prerequisites}
        \label{fig:ex3}
    \end{figure}

    \begin{figure}[t]
        \centering
        \includegraphics[width=\linewidth]{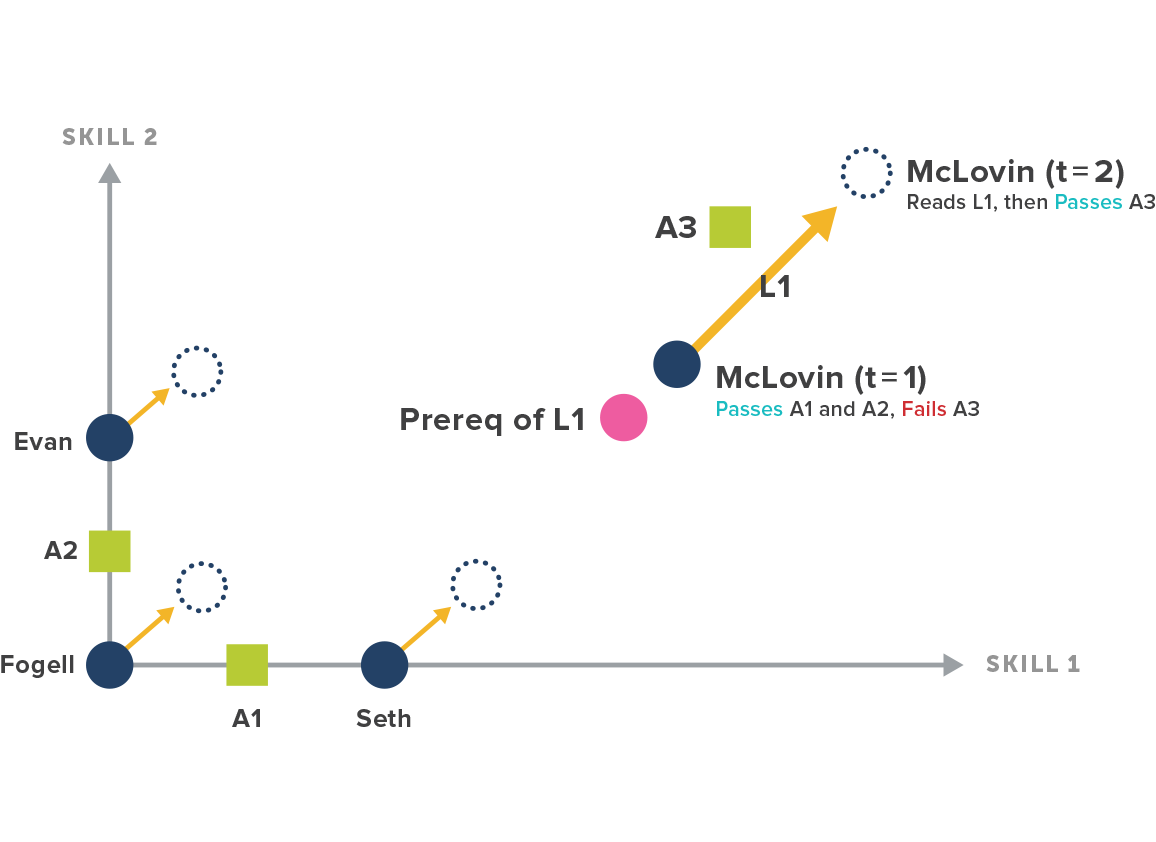}
        \caption{A two-dimensional embedding with lessons and prerequisites}
        \label{fig:ex4}
    \end{figure}

Figure \ref{fig:ex1} demonstrates an extremely simple embedding. The key observation here is that the model recovered positive skill gains for lesson L1, and ``correctly'' arranged Alice and assessment A1 in the latent space. Initially, Alice fails A1, so her skill level is behind the requirements of A1. After completing L1, Alice passes A1, indicating that her skill level has probably improved past the requirements of A1. Note that this scenario could have been explained with only one latent skill.

Figure \ref{fig:ex2} depicts a two-dimensional embedding, where an intransitivity in assessment results requires more than one latent skill to explain. The key observation here is that the assessments are embedded on two different axes, meaning they require two completely independent skills. This makes sense, since student results on A1 are uncorrelated with results on A2. Fogell fails both assessments, so his skill levels are behind the requirements for A1 and A2. McLovin passes both assessments, so his skill levels are beyond the requirements for A1 and A2. Evan and Seth are each able to pass one assessment but not the other. Since the assessments have independent requirements, this implies that Evan and Seth have independent skill sets (i.e. Evan has enough of skill 2 to pass A2 but not enough of skill 1 to pass A1, and Seth has enough of skill 1 to pass A1 but not enough of skill 2 to pass A2).

In Figure \ref{fig:ex3}, we replicate the setting in Figure \ref{fig:ex2}, then add two new students Slater and Michaels, and two new lesson modules L1 and L2. Slater is initially identical to Evan, while Michaels is initially identical to Seth. Slater reads lesson L1, then passes assessments A1 and A2. Michaels reads lesson L2, then passes assessments A1 and A2. The key observation here is that the skill gain vectors recovered for the two lesson modules are orthogonal, meaning they help students satisfy completely independent skill requirements. This makes sense, since initially Slater was lacking in Skill 1 while Michaels was lacking in Skill 2, but after completing their lessons they passed their assessments, showing that they gained from their respective lessons what they were lacking initially.

In Figure \ref{fig:ex4}, we replicate the setting in Figure \ref{fig:ex2}, then add a new assessment module A3 and a new lesson module L1. All students initially fail assessment A3, then read lesson L1, after which McLovin passes A3 while everyone else still fails A3. The key observation here is that McLovin is the only student who initially satisfies the prerequisites for L1, so he is the only student who realizes significant gains from taking L1.

\section{Experiments on Online Course Data}

We use data processed by Knewton, an adaptive learning technology company. Knewton's infrastructure uses student-content access traces to generate personalized recommendations and activity analytics for partner organizations with online learning products. The data describes interactions between college students and two science textbooks. The Book A data set was collected from 869 classrooms from January 1, 2014 through June 1, 2014. It contains 834,811 interactions, 3,471 students, 3,374 lessons, 3,480 assessments, and an average assessment pass rate of 0.712. The paths that students take are biased by direction from instructors, a recommender system, and the sequence of chapters in the textbook. The Book B data set was collected from 1,070 classrooms from January 1, 2014 through June 1, 2014. It contains 1,349,541 interactions, 3,563 students, 3,843 lessons, 3,807 assessments, and an average assessment pass rate of 0.693.

Both data sets are filtered to eliminate students with fewer than five lesson interactions and content modules with fewer than five student interactions. To avoid spam interactions and focus on the outcomes of initial student attempts, we only consider the first interaction between a student and an assessment (subsequent interactions between student and assessment are ignored).

\subsection{Assessment Result Prediction} \label{assessment-result-prediction}

\begin{table*}[t]
\caption{Test AUC, validation AUC, and standard error of validation AUC for variations of the embedding model and benchmark IRT models.}
\label{all-results}
\begin{center}
\arrayrulecolor{lightgray}
\setlength\tabcolsep{7mm}
\begin{tabular}{ l | l | l | l | l | l | l | l }
    & \multicolumn{3}{c|}{Model} & \multicolumn{2}{c|}{Book A} & \multicolumn{2}{c}{Book B} \\
    & $\vec{\ell}$ & $\vec{q}$ & $\gamma$ & Test & Validation & Test & Validation \\  \hline
    \textbf{1} & N & N & N & 0.673 & $0.614 \pm 0.015$ & 0.614 & $0.644 \pm 0.015$ \\ \hline
    \textbf{2} & N & N & Y & 0.818 & $0.753 \pm 0.020$ & 0.788 & $0.821 \pm 0.021$ \\ \hline
    \textbf{3} & Y & N & N & 0.692 & $0.624 \pm 0.019$ & 0.630 & $0.662 \pm 0.023$ \\ \hline
    \textbf{4} & Y & N & Y & 0.798 & $0.761 \pm 0.016$ & 0.775 & $0.808 \pm 0.020$ \\ \hline
    \textbf{5} & Y & Y & N & 0.724 & $0.625 \pm 0.021$ & 0.629 & $0.643 \pm 0.018$ \\ \hline
    \rowcolor{yellow}
    \textbf{6} & Y & Y & Y & 0.811 & $0.756 \pm 0.018$ & 0.785 & $0.823 \pm 0.021$ \\ \hline
    \textbf{7} & \multicolumn{3}{c|}{1PL IRT} & 0.812 & $0.761 \pm 0.016$ & 0.778 &  $0.812 \pm 0.019$ \\ \hline
    \textbf{8} & \multicolumn{3}{c|}{2PL IRT} & 0.780 & $0.708 \pm 0.011$ & 0.686 & $0.690 \pm 0.022$ \\ \hline
    \textbf{9} & \multicolumn{3}{c|}{2D MIRT} & 0.817 & $0.732 \pm 0.012$ & 0.776 & $0.796 \pm 0.018$ \\
\end{tabular}
\end{center}
\end{table*}

We evaluate the embedding model on the task of predicting results of held-out assessment interactions, and compare it to three benchmark models: the one- and two-parameter logistic item response theory models, and a two-dimensional item response theory model. The 1PL IRT model, also known as the Rasch model, has the following assessment pass likelihood: $\mathbb{P}[R=1] = \phi(\theta_i - \beta_j)$ for student $i$ and item $j$, where $\theta$ is student proficiency and $\beta$ is item difficulty, and $\phi$ is the logistic link function \cite{rasch1993probabilistic}. The 2PL model extends the likelihood as follows: $\mathbb{P}[R=1] = \phi(\alpha_j(\theta_i - \beta_j))$, where $\alpha$ is the item discriminability \cite{linden1997handbook}. The 2D MIRT model, which is a multi-dimensional generalization of 2PL, has the following pass likelihood: $\mathbb{P}[R=1] = \phi(\vec{u_i} \cdot \vec{v_j} + \mu_j)$, where $\vec{u}$ are the student factors, $\vec{v}$ are the item factors, and $\mu$ is the item offset \cite{m.d.reckase2009}. Note that we have not explicitly included Bayesian Knowledge Tracing as a benchmark model since it requires content modules to be annotated with concept tags, while the Latent Skill Embedding does not.

\begin{figure}[t]
\centering
\includegraphics[width=\linewidth]{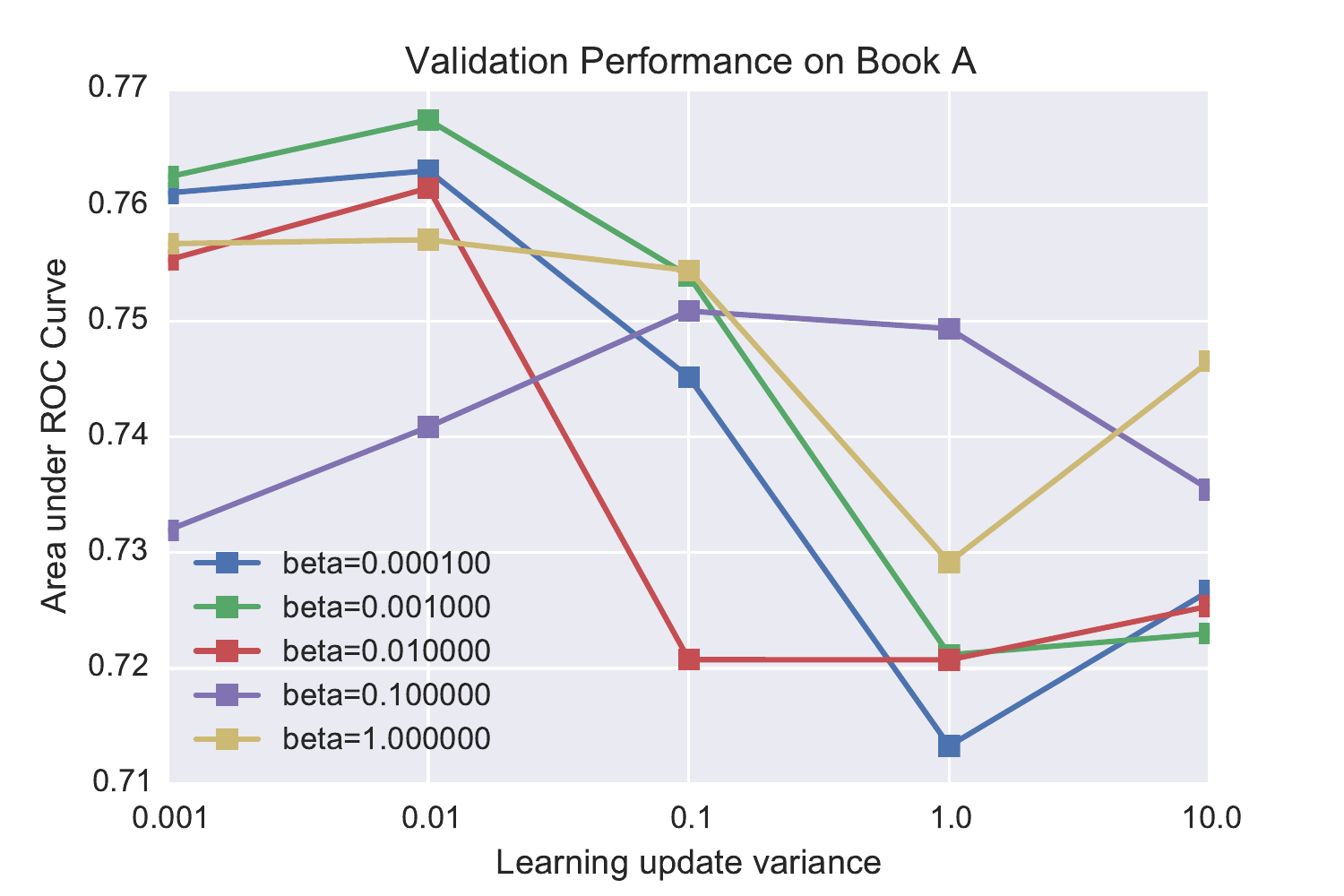}
\caption{We explore the parameter space of the two-dimensional embedding with prerequisites and bias terms by doing a grid search on $(\sigma^2, \beta)$.}
\label{fig:betaluv-grid}
\end{figure}

We use ten-fold cross-validation to select the regularization parameter $\beta$ and learning update variance $\sigma^2$ for the embedding model (see Figure \ref{fig:betaluv-grid} for the exploration on Book A), as well as regularization parameters for the benchmark IRT models. On each fold, we train on the full histories of 90\% of students and the truncated histories of 10\% of students, and validate on the assessment interactions immediately following the truncated histories. Truncations are made just before the last assessment interactions for each student (maximizing the size of the training set). We have also examined the effect of randomizing the truncation in student histories, and find no substantial changes to our results.

After selecting model hyperparameters using cross-validation, we evaluate the models on a held-out test set of students (20\% of the students in the complete data set) that was not visible during the earlier parameter selection phase. The same truncation method is used for evaluation on the test set.

Our performance metric is area under the ROC curve (AUC), which measures the discriminative ability of a binary classifier that assigns probabilities to class membership.

\paragraph{Lesion Analysis}
To gain insight into which components of the embedding model contribute most to its predictive power, we conduct a lesion analysis. For the sake of simplicity, we restrict ourselves to using a two-dimensional embedding (later, we describe the effect of varying the embedding dimension $d$). We start with an embedding model that ignores lesson interactions and does not use bias terms. We then gradually add components to the embedding model to examine their effects on prediction AUC. Specifically, we evaluate embeddings with and without lesson parameters $\vec{\ell}$, prerequisite parameters $\vec{q}$ for lessons, and bias terms $\gamma$. Each variant of the model corresponds to a row in Table \ref{all-results}.

From these results, we observe the following: including bias terms in the assessment result likelihood (Equation \ref{eq:result-likelihood}) gives a large and statistically significant performance gain ($p<0.0003$ for the standard t-test comparing validation AUCs of row 5 vs. 6 on Book A); an embedding with lesson prerequisites and bias terms performs comparably to the best benchmark IRT model.

\paragraph{Effect of Data Heterogeneity}
One issue that may have affected the findings is the biased nature of student paths, which has been discussed by \cite{gonzalez2014general}. In the data, we observe that student paths are heavily directed along common routes through modules. We conjecture that this bias dulls the effect of modeling lesson prerequisites in the embedding, and causes the inclusion of bias terms to give a large performance boost. Most students attack a module with the same background knowledge, so an embedding that captures the variation in students who work on the same module is not as valuable.

\begin{figure}[t]
\centering
\includegraphics[width=\linewidth,trim=8 0 35 0,clip]{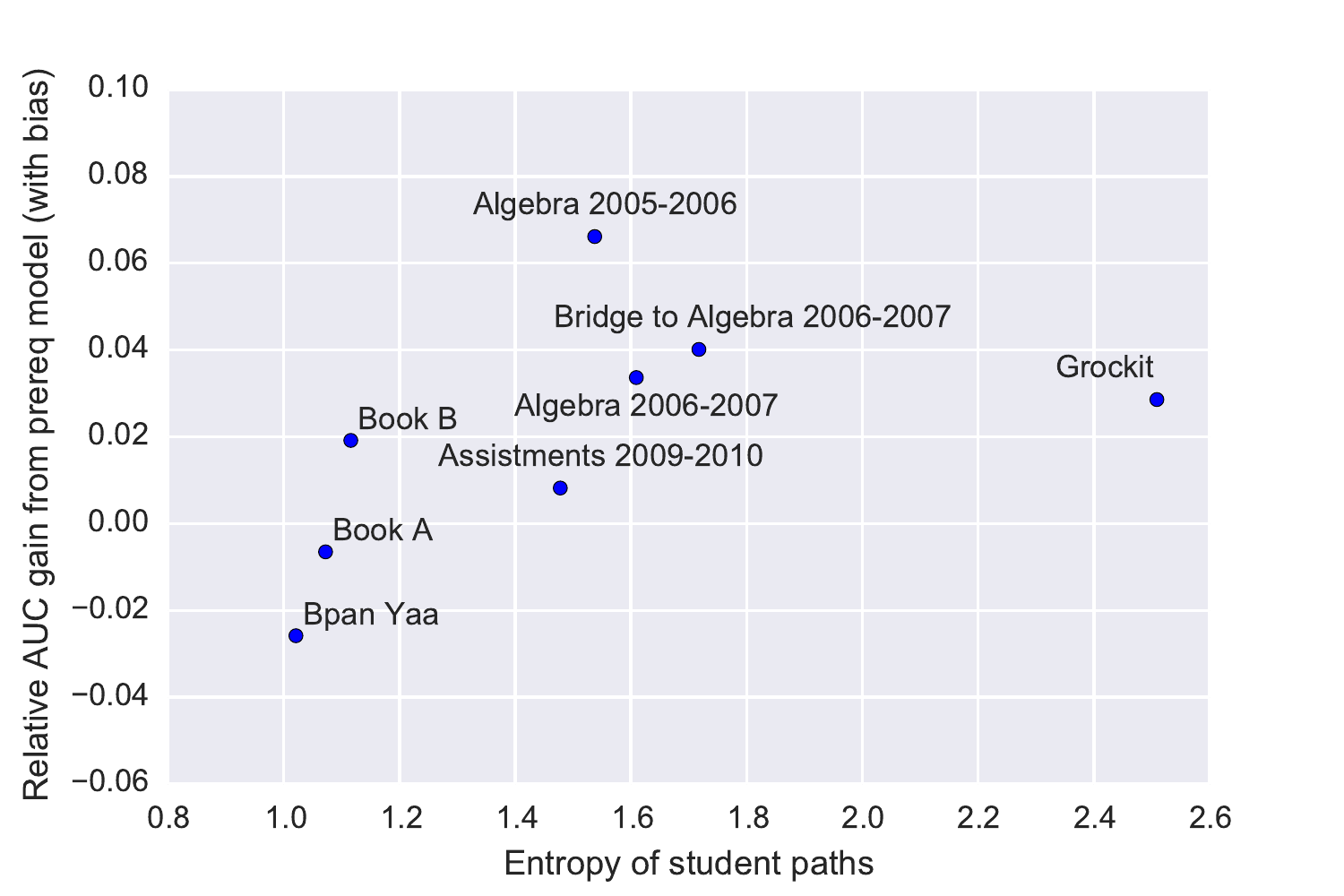}
\caption{The gain from including lesson prerequisites in the embedding seems to depend on the entropy of student paths in the data set.}
\label{fig:gain-vs-entropy}
\end{figure}

In a regime where students who work on a module come from a variety of skill backgrounds, our model that includes lesson prerequisites may further improve results. Preliminary evidence for this is presented in Figure \ref{fig:gain-vs-entropy}, where we re-create our analysis of the two Knewton data sets on several public data sets of student interactions and a private data set from an online language learning game \cite{kddcup10, feng2009addressing, andersen2014automatic, grockit}. After conducting a lesion analysis for each data set, we compute the relative difference between the validation AUC for a two-dimensional embedding \emph{with} lesson prerequisites and the validation AUC for a two-dimensional embedding \emph{without} lesson prerequisites. We also measure the ``entropy'' of student paths by assuming that student paths can be modeled as Markov chains and computing the entropy of the transition probability matrix for each data set. A relationship exists between student path entropy and the relative AUC gain from using lesson prerequisites, providing some evidence to support the hypothesis that modeling lesson prerequisites is more beneficial in a setting with larger variation in student paths.

\paragraph{Effect of Embedding Dimension}
In other experiments, we explored the parameter space of the embedding model by varying the regularization constant $\beta$ and embedding dimension $d$. Not explicitly shown are the results for changing $d$. In summary, we find that increasing embedding dimension $d$ substantially improved performance for embedding models without bias terms, but that it has little effect on performance for embeddings with bias terms. The former is expected, since the embedding itself must be used to model general student passing ability and general assessment difficulty.

\begin{figure}[t]
\centering
\includegraphics[width=\linewidth,trim=8 0 32 0,clip]{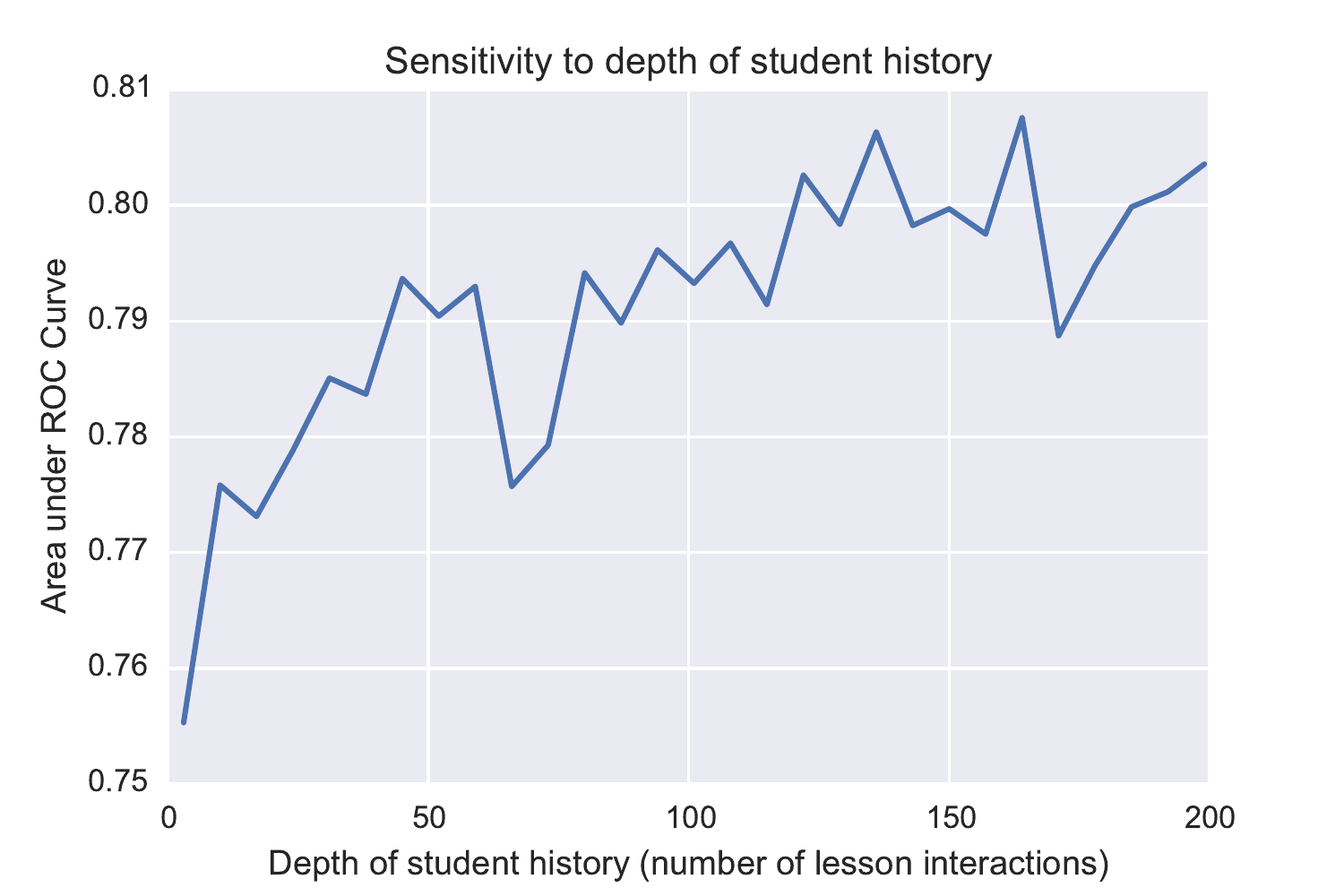}
\caption{Sensitivity of validation AUC to the ``depth" of a student's history (from $t=T-depth$ to $t=T$). A student's recent history is most helpful for predicting assessment results, which we observe in the plateauing of the curve as we gradually include interactions from the students far past.}
\label{fig:depth-sensitivity}
\end{figure}

\begin{figure}[t]
\centering
\includegraphics[width=\linewidth,trim=8 0 32 0,clip]{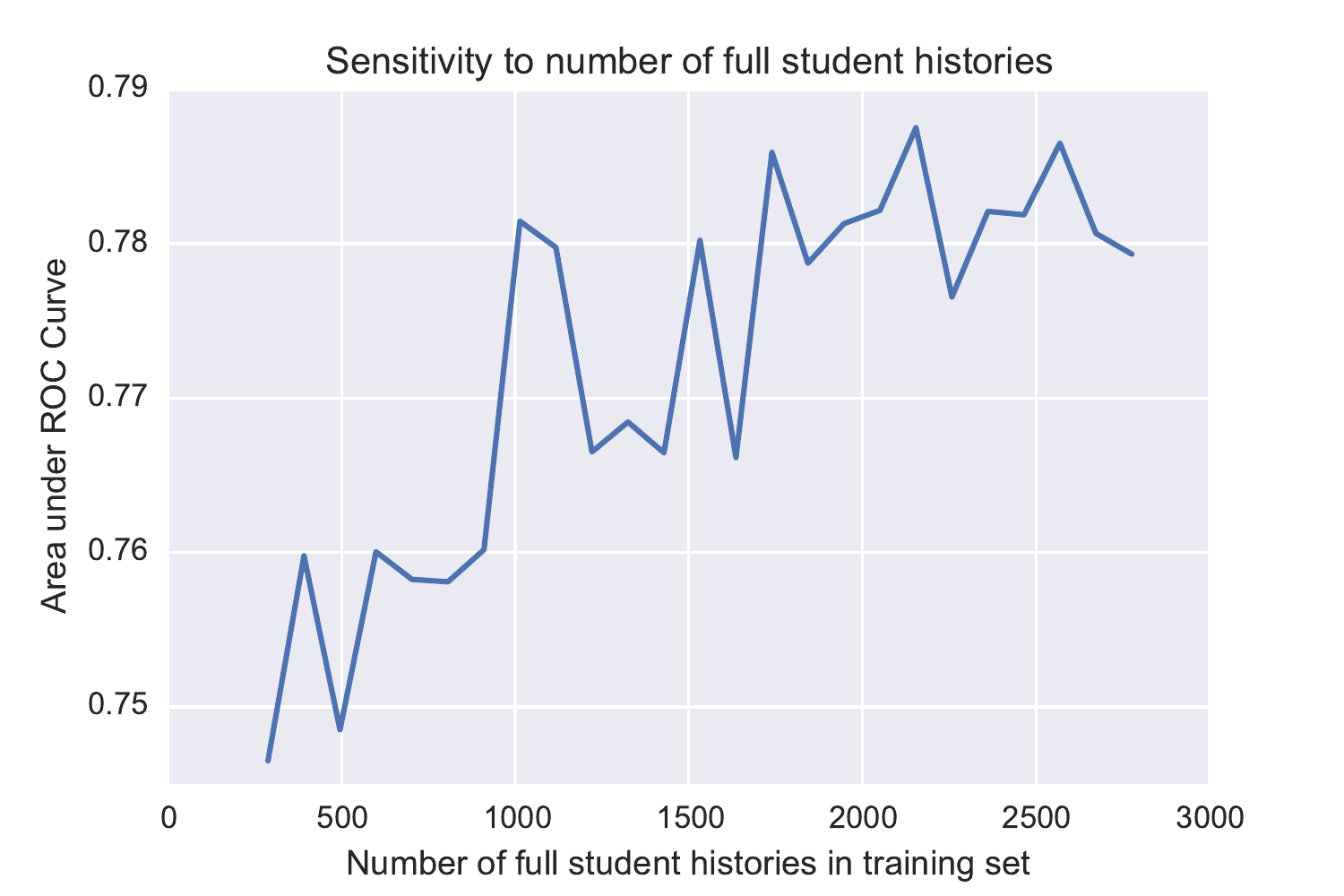}
\caption{Sensitivity of validation AUC to the number of full student histories in the training set. The number of full histories affects the quality of module embeddings, and thus has a strong effect on performance.}
\label{fig:tset-sensitivity}
\end{figure}

\begin{figure}[t]
\centering
\includegraphics[width=\linewidth,trim=8 0 35 0,clip]{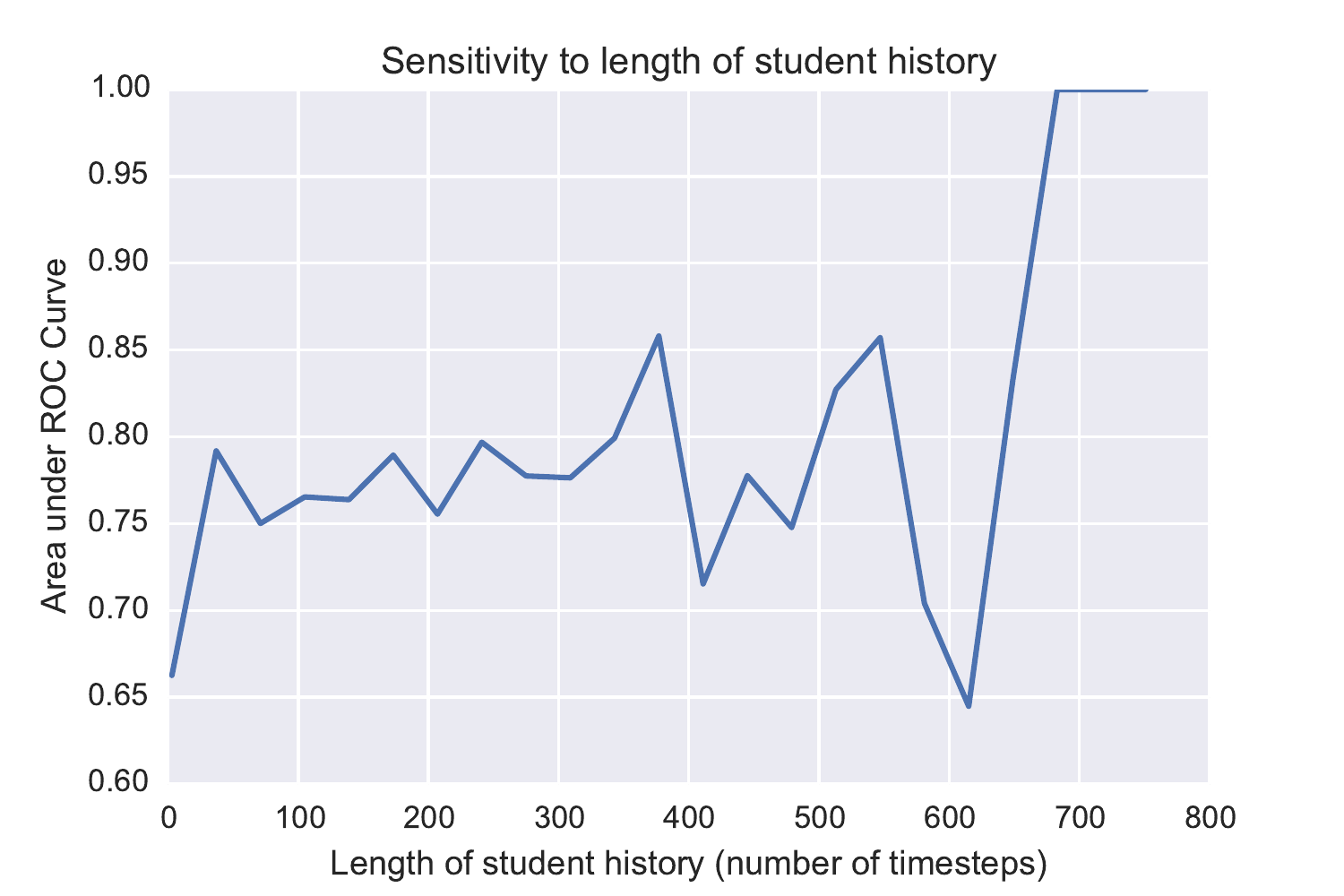}
\caption{Sensitivity of validation AUC to the length of a student history (from $t=0$ to $t=T$). There is a very weak and noisy relationship between history length and performance.}
\label{fig:length-sensitivity}
\end{figure}

\begin{figure}[t]
\centering
\includegraphics[width=\linewidth,trim=8 0 35 0,clip]{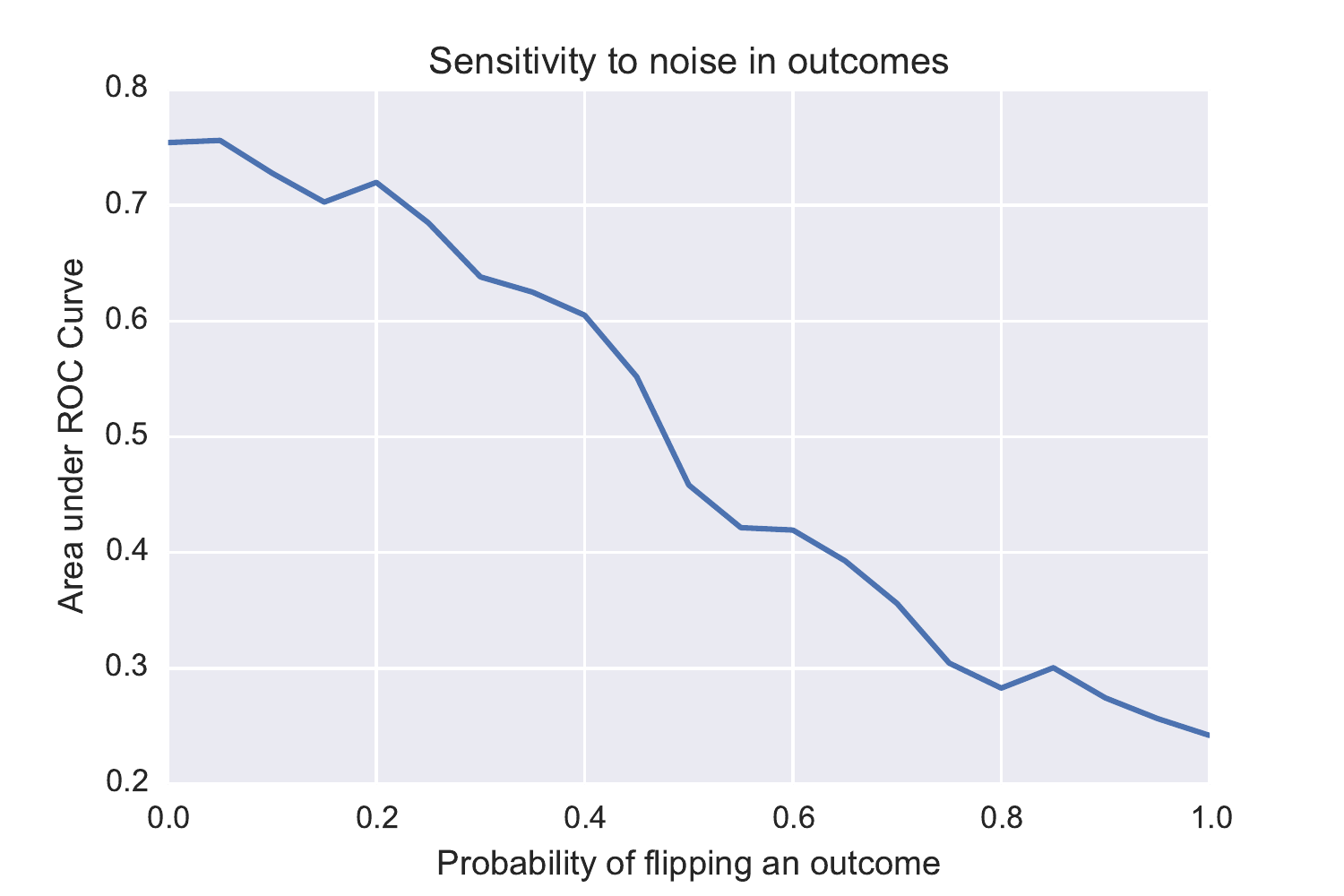}
\caption{Sensitivity of validation AUC to noisy assessment results in the training set. A small amount of noise is acceptable, but performance quickly deteriorates as the training set stops ``agreeing" with the validation set and eventually (when AUC drops below 0.5) starts biasing the model so that it performs worse than a random coin flip. In practice, we anticipate a very small amount of noise.}
\label{fig:noise-sensitivity}
\end{figure}

\paragraph{Sensitivity Analysis}
We perform several sensitivity analyses on Book A and observe the following: prediction AUC is most affected by a student's recent history (see Figure \ref{fig:depth-sensitivity}); the number of full student histories in the training set has a strong effect on prediction AUC, via the quality of module embeddings (see Figure \ref{fig:tset-sensitivity}); prediction AUC decays when assessment results in the training set are noisy (see Figure \ref{fig:noise-sensitivity}); the length of a student's history is weakly related to prediction AUC (see Figure \ref{fig:length-sensitivity}). These findings lead to two key qualitative insights regarding model performance: (1) for a course offered regularly (e.g., over several semesters), the model will improve steadily as log data is collected from students who complete the course, and (2) the model performs best when assessments are crafted to test specific skills and minimize noise in outcomes.


\subsection{Lesson Sequence Discrimination}

The ability to predict future performance of students on assessments, while a useful metric for evaluating the learned embedding, does not address the more important task of adaptive tutoring via customized lesson sequence recommendation. We introduce a surrogate task for evaluating the sequence recommendation performance of the model based entirely on the observational data of student interactions, by assessing the model's ability to recommend ``productive'' paths amongst several alternatives.

\begin{figure}[t]
\centering
\includegraphics[height=0.5\linewidth]{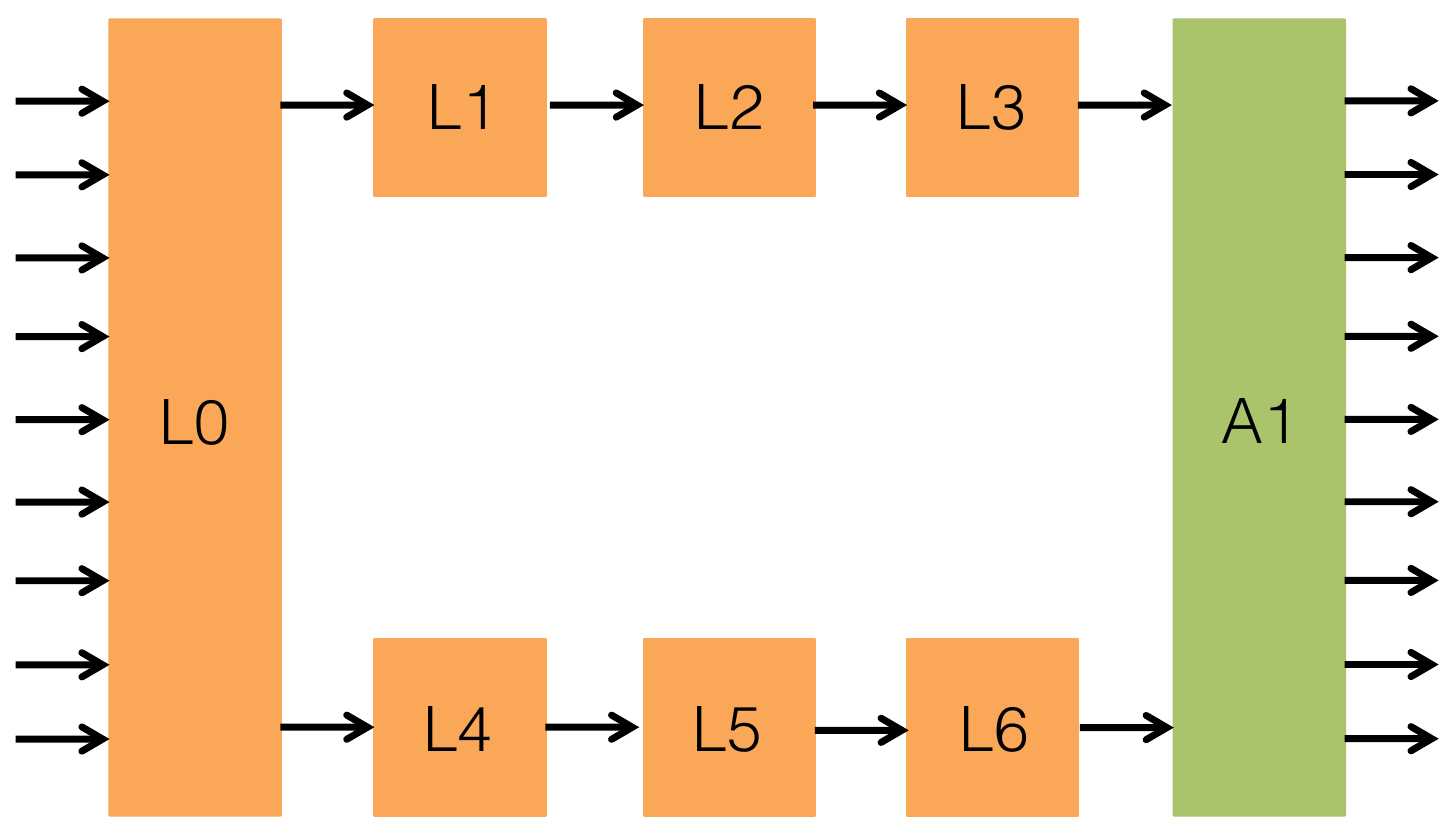}
\caption{A schematic diagram of a \emph{bubble}, where a group of students converges on a lesson, splits off into two different lesson sequences, then converges on the same assessment.}
\label{fig:bubble-schematic}
\end{figure}

\paragraph{Bubbles as Experimental Evidence} The size of the data set creates a unique opportunity to leverage the variability in learning paths to simulate the setting of a controlled experiment. For this evaluation, we use a larger version of the Book A data set, containing 14,707 students and 14,327 content modules. We find that the data contains many instances of student paths that share the same lesson module at the beginning and the same assessment module at the end, but contain different lessons along the way. We call these instances \emph{bubbles}, for example see Figure \ref{fig:bubble-schematic}, which present themselves as a sort of experimental evidence on the relative merits of two different learning progressions. We can thus use these \emph{bubbles} to evaluate the ability of an embedding to recommend a learning sequence that leads to success, as measured by the relative performance of students who take the recommended vs. the not-recommended path to the assessment module at the end of the \emph{bubble}.

We use the full histories of 70\% of students to embed lesson and assessment modules, then train on the histories of held-out students up to the beginning of a \emph{bubble}. The lesson sequence for a student is then simulated over the initial student embedding, using the learning update (Equation \ref{eq:update-with-prereqs}) to compute an expected student embedding at the end of the \emph{bubble} (which can be used to predict the passing likelihood for the final assessment using Equation \ref{eq:result-likelihood}). The path that leads the student to a higher pass likelihood on the final assessment is the ``recommended'' path. Our performance measure is $\mathbb{E}\left[\frac{\mathbb{E}[R'] - \mathbb{E}[R]}{\mathbb{E}[R]}\right]$, where $R' \in \{0, 1\}$ is the outcome at the end of the recommended path and $R \in \{0, 1\}$ is the outcome at the end of the other path ($0$ is failing and $1$ is passing). This measure can be interpreted as ``expected gain'' (averaged over many \emph{bubbles}) from taking recommended paths, or how ``successful'' the paths recommended by the model are when compared to the alternative.

\paragraph{Propensity Score Matching} This observational study is potentially confounded by many hidden variables. For example, it may be that one group of students systematically takes recommended paths while another group of students does not, leading to results at the end of a \emph{bubble} that are mostly dictated by the teachers directing the groups, or other student-specific hidden factors, rather than path quality. To best approximate the settings of a randomized controlled trial in our observational study, we use the standard \emph{propensity score matching} approach for de-biasing observational data \cite{rosenbaum1983central, caliendo2008some}. The key idea behind \emph{propensity score matching} is to subset the observed data in a way that balances the distribution of the features (``hidden variables'') describing subjects in the two conditions, as it would be expected in a randomized experiment. The validity of any conclusion drawn from the observational data de-biased in this way hinges on the assumption that all confounding variables that determine self-selection have been accounted for in the features prior to matching. In this study, we hypothesize that the set of all lesson modules and assessment modules (with outcomes) that the learner attempted throughout his or her duration in the online system is sufficient to compensate for any self-selection in the taken learning paths. Formally, we represent learners in a feature space $X$ such that $X_{ij} \in \{-1, 0, 1\}$, where $X_{ij} = 1$ if student $i$ passed module $j$ (lessons are always ``passed''), $X_{ij} = 0$ if student $i$ has not completed module $j$, and $X_{ij} = -1$ if student $i$ failed module $j$.

\begin{figure}[t]
\centering
\includegraphics[width=\linewidth,trim=30 0 50 0,clip]{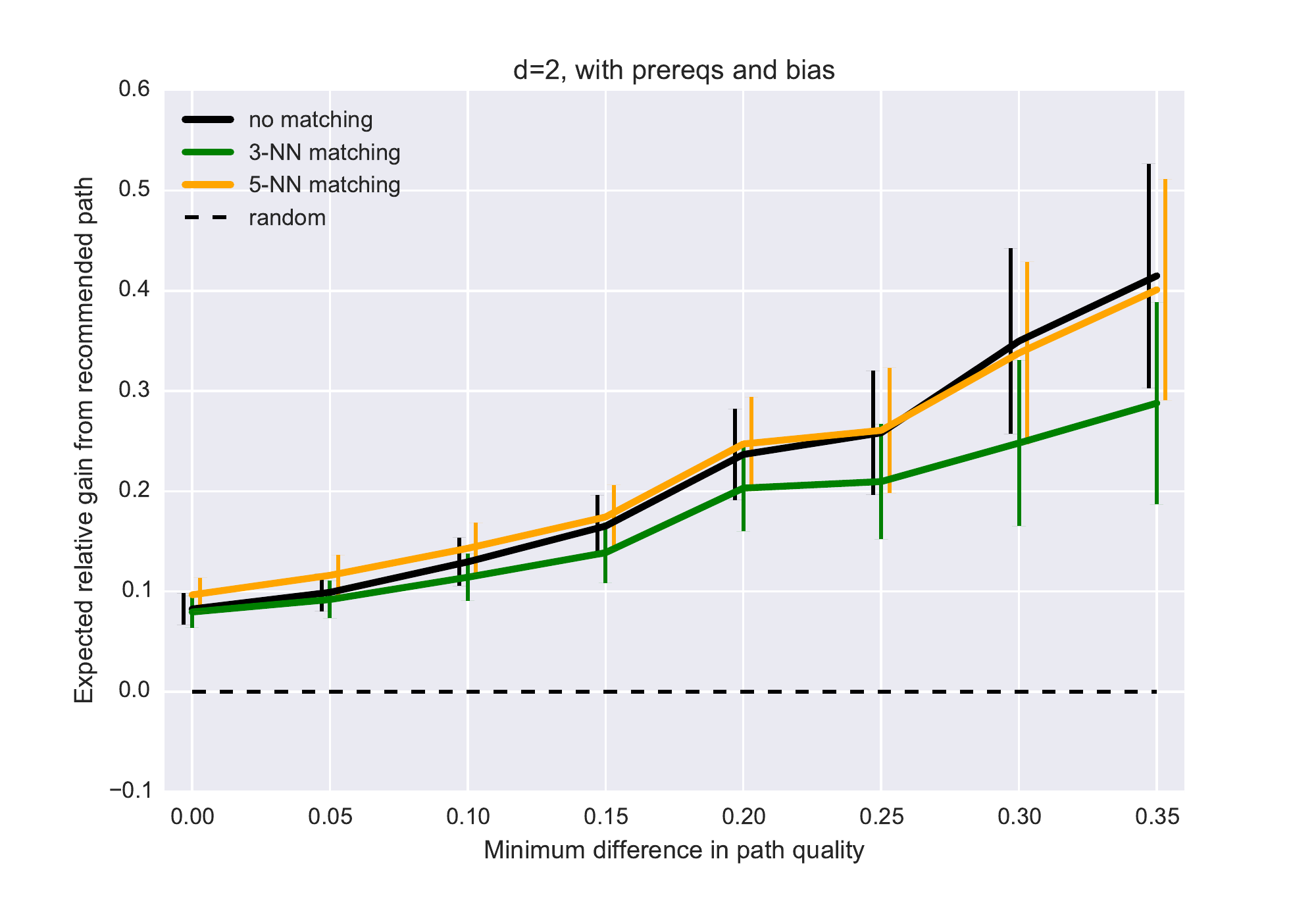}
\caption{The \emph{x}-axis represents a threshold on absolute difference between pass rates of the two \emph{bubble} paths. \emph{Bubbles} are filtered to meet the following criteria: at least ten students take each branch, each branch must contain at least two lessons, and both branches must contain the same number of lessons. The error bars represent standard error, and their x-coordinates are slightly perturbed so the error bars for different curves can be distinguished.}
\label{fig:gain-vs-armdiff}
\end{figure}

We use PCA to map $X$ to a low-dimensional feature space where students are described by 1,000 features, which capture 80\% of the variance in the original 14,327 features. A logistic regression model with $L_2$ regularization is used to estimate the probability of a student following the recommended branch of a \emph{bubble}, i.e. the \emph{propensity score}, given the student features (the regularization constant is selected using cross-validation to maximize average log-likelihood on held-out students). Within each \emph{bubble}, students who took their recommended branch are matched with their nearest neighbors (by absolute difference in \emph{propensity scores}) from the group of students who did not take their recommended branch. Matching is done with replacement (so the same student can be selected as a nearest neighbor multiple times) to improve matching quality, trading off bias for variance. Multiple nearest neighbors can be matched (we examine the effect of varying $k$), trading off variance for bias.

\paragraph{Results} Figure \ref{fig:gain-vs-armdiff} shows the results of the experiment, showing by how much students gain by following the path recommended by our embedding. We use the same embedding configuration as in row 6 of Table \ref{all-results}, which uses prerequisites and bias terms in a two-dimensional embedding model with lesson. Naturally, our evaluation metric of gain in the pass rate from following a recommended path would depend strongly on the relative merits of the recommended and alternative paths. We therefore plot the gain that the recommended path achieves in relation to the difference in path quality, as measured by the absolute difference in pass rates between the two paths. Figure \ref{fig:gain-vs-armdiff} shows that the model generally able to recommend more successful paths, and this finding is robust to the choice of nearest neighbors $k$ used during propensity matching. As expected, the effect of the system recommendation is larger when there is a significant difference between the quality of the two paths.

\section{Conclusions} \label{conclusion}

We presented a general model that learns a representation of student knowledge and educational content that can be used for personalized instruction. The key idea lies in using a multi-dimensional embedding to capture the dynamics of learning and testing. Using a large-scale data set collected in real-world classrooms, we (1) demonstrate the ability of the model to successfully predict learning outcomes and (2) introduce an offline methodology as a proxy for assessing the ability of the model to recommend personalized learning paths. We show that our model is able to successfully discriminate between personalized learning paths that lead to mastery and failure.

An implementation of the Latent Skill Embedding and the IPython notebooks used to conduct experiments are available online at \url{http://siddharth.io/lentil}.

\section{Acknowledgements}

We would like to thank members of Knewton for valuable feedback on this work. This research was funded in part by the National Science Foundation under Awards IIS-1247637, IIS-1217686, and IIS-1513692, a grant from the John Templeton Foundation provided through the Metaknowledge Network at the University of Chicago and through the Cornell Presidential Research Scholars program.

%
\bibliographystyle{abbrv}
\bibliography{manuscript}  
%
%

\end{document}